\documentclass[10pt,twocolumn,letterpaper]{article}
\usepackage[pagenumbers]{cvpr}

\usepackage{stfloats}
\usepackage{array}
\definecolor{linkblue}{rgb}{0.21,0.49,0.74}
\usepackage[pagebackref,breaklinks,colorlinks,allcolors=linkblue]{hyperref}
\title{FMSNO: Scaling Surface Neural Operators by Learned Flow Matching Reparametrisation}
\author{Vasilii Liamin \quad Oleg Bakhteev \quad Vladimir Vanovskiy}
\hypersetup{pdfauthor={Vasilii Liamin, Oleg Bakhteev, Vladimir Vanovskiy},pdftitle={FMSNO: Scaling Surface Neural Operators by Learned Flow Matching Reparametrisation}}

\begin{document}
\maketitle
\begin{figure*}[!b]
  \centering
  \includegraphics[width=\textwidth]{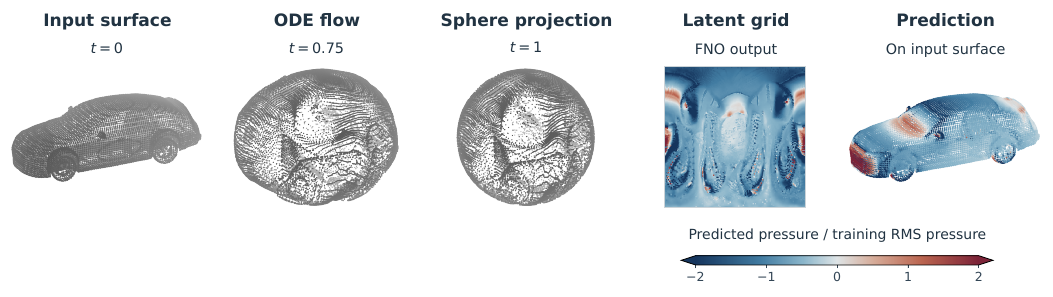}
  \caption{Surface pressure prediction on a DrivAerNet++ test car. The first three panels show the geometry along the learned flow and after projection onto the sphere. We splat geometric features onto the latent grid, where the FNO predicts pressure (fourth panel), and interpolate this field back to the input surface (fifth panel). Both pressure panels use the same colour scale.}
  \label{fig:drivaer-pipeline}
\end{figure*}
\begin{abstract}
We propose the Flow Matching Surface Neural Operator (FMSNO), which combines learning a shared map from irregular surface meshes to a regular two-dimensional grid on the sphere and training a neural operator on the regular grid.
Constructing a separate map for each new geometry can dominate inference time on large meshes.
We initially pretrain the mapping by flow matching using precomputed optimal transport maps as targets and then continue its training using combined geometry and CFD objectives.
We apply it to unseen geometries in a single forward pass.
The learned flow deforms the surface points, and we project its endpoints onto the sphere.
We then splat geometric features onto the spherical grid, apply a neural operator (we tested both Fourier and Spherical Fourier Neural Operators), and interpolate its output back to the surface.
For a mesh with $N$ points, evaluating the map costs $O(N)$, and the full prediction costs $O(N\log N)$ with the FNO backbone.
We evaluate surface pressure prediction on ShapeNet-Car, AhmedML, and DrivAerNet++, and drag coefficient prediction on DrivAerNet++.
Across the three pressure benchmarks, our accuracy is on par with that of the most accurate baseline.
On a surface with one million points, our method achieves a $354\times$ end-to-end inference speedup over that baseline while keeping the same memory usage asymptotics of $O(N)$.
\end{abstract}

\section{Introduction}
\label{sec:intro}

Simulating physical fields on complex three-dimensional geometries is a recurring bottleneck in
scientific computing \citep{slotnick2014cfdvision,azizzadenesheli2024neuralop}. Spectral solvers
are efficient on regular domains \citep{gottlieb1977spectral,boyd2001spectral}, and on an arbitrary
shape the whole solve is repeated from scratch whenever the geometry changes. The difficulty is common to fluid dynamics
\citep{slotnick2014cfdvision}, solid mechanics \citep{bazilevs2010tsplines} and geoscience
\citep{komatitsch1999sem}, and it is acute in design loops, which evaluate many shapes rather
than one \citep{li2023geofno,li2023gino}. A single steady-state CFD solution around one object costs about two hundred CPU-hours
\citep{drivaernet}, or more than ten hours even on GPUs \citep{li2023gino,korzun2022des}, and shape
optimisation requires thousands of them.

We train and test on different shapes from the same family, such as vehicle bodies.
Each input is an irregular surface mesh with its own vertex count and locations, and the output is a field on those vertices.

Fourier neural operators (FNOs) take advantage of fast Fourier transforms on regular
grids~\citep{li2021fno,li2023geofno}. To apply one to an irregular surface mesh, we need a map
that places the surface features on such a grid and lets us evaluate the predicted field back
on the mesh. Computing a new map for every geometry can dominate inference time on large meshes.

We learn a surface-to-sphere map that can be applied directly to a new geometry.
Our continuous-flow reparametrisation uses one velocity field shared by all shapes.
We integrate this field from each surface point, project the endpoints onto the sphere, and
splat the surface features onto an equiangular latitude--longitude grid. An FNO or its spherical
variant (SFNO)~\citep{bonev2023sfno} acts on this grid, and bilinear interpolation returns the predicted field to
the surface points. We focus on automotive aerodynamics because vehicle designers compare many candidate body shapes, estimating the surface pressure distribution and drag coefficient for each~\citep{li2023gino,drivaernetpp}.
We evaluate the method on three surface-pressure datasets and one drag task.
Figure~\ref{fig:drivaer-pipeline} shows each stage on a held-out DrivAerNet++ geometry.

\paragraph{Contributions.}
\begin{enumerate}\itemsep2pt
  \item \textbf{Method.} For a new surface with $N$ points, evaluating the learned map takes
        $O(N)$ time and the full prediction takes $O(N\log N)$ time with the FNO backbone. On a new synthetic surface
        with one million points, we measure a $354\times$ end-to-end inference speedup over
        OTNO~\citep{li2025otno}, a state-of-the-art neural operator that computes a separate
        map for each geometry (Fig.~\ref{fig:inference_scaling}).
  \item \textbf{Results.}\footnote{We will release the code with the camera-ready version.} Our pressure prediction accuracy is on par with this state-of-the-art
        baseline across the three datasets. FMSNO achieves lower pressure error than our reproduction of OTNO on ShapeNet-Car and
        AhmedML. On DrivAerNet++, our FNO achieves a drag $R^2$ of $0.883$, close to OTNO's $0.887$ (\S\ref{sec:exp}).
  \item \textbf{Analysis.} We quantify how much bilinear splatting and interpolation distort the
        reference pressure field and show how joint training reduces this distortion (\S\ref{sec:analysis}).
\end{enumerate}

\section{Related work}
\label{sec:related}

\paragraph{Surrogates for fields on geometries.}
Learned surrogates replace each solve with a forward pass. They differ in what they predict.
Volumetric surrogates reconstruct the field in the space around the body. They must discretise that volume, and their cost grows with its resolution. Resolving fine detail at the boundary requires a fine discretisation near it.
Surface surrogates predict fields on the boundary itself, typically the pressure~\citep{li2023gino,li2025otno,drivaernet}.
Integrating pressure and wall shear stress over the surface gives forces such as drag and lift.

\paragraph{Neural operators on irregular geometries.}
Fourier neural operators~\citep{li2021fno} act on a regular grid. There are several ways to map a surface onto a reference domain that carries such a grid, and they differ above all in how much work a new shape costs.
Some methods learn the map once and reuse it for every geometry. Others rebuild it for each one.

Geo-FNO~\citep{li2023geofno} distinguishes two cases. When the input mesh is structured, its indexing already induces a coordinate map, and the method reduces to a standard FNO on that mesh. When the mesh is unstructured, the map is a network trained end-to-end with the operator. The network can condition the displacement at each point on a fixed-length shape code shared across the points of that geometry. In the original formulation, the reference domain has the same dimension as the physical domain.

GINO~\citep{li2023gino} sits between the two families. Its local graph kernel is learned once, but its neighbourhoods are rebuilt for each shape.
In its decoder-only model for surface pressure, a signed distance function represents the vehicle geometry on a regular three-dimensional grid. An FNO acts on this grid, and a graph neural operator decodes pressure at the queried surface points.
The cost of the FNO therefore depends on the resolution of the volume grid. In the authors' Ahmed-body experiment, the inlet speed is an additional scalar boundary condition.

OTNO~\citep{li2025otno} builds the map separately for each shape by solving an optimal transport problem between the surface points and a two-dimensional latent grid.
With a single surface point assigned to each latent node, its encoder copies that point's features directly to the grid. The plan is a property of one mesh and must be solved again for every new geometry at inference, at $O(N^2)$ cost for a grid of $\Theta(N)$ nodes and a fixed number of Sinkhorn iterations.

\paragraph{Transformer-based solvers.}
Transformer encoders can operate directly on irregular point sets. Full self-attention over $N$ points has quadratic cost, motivating architectures that aggregate points before computing global interactions.
Transolver~\citep{wu2024transolver} learns soft assignments of mesh points to a set of slices.
It pools the point features in each slice into a token, applies attention across the tokens, and distributes the updated features back to the mesh points.
For a fixed number of slices, this computation scales linearly with the number of points.
MSPT~\citep{curvo2026mspt} combines attention within spatial patches with global attention across patch representations to predict aerodynamic fields, including surface pressure on ShapeNet-Car and AhmedML.

\paragraph{Implicit neural representations.}
An alternative is to represent a field as a continuous function of spatial coordinates~\citep{yin2023dino,chen2023insr}.
The field can then be evaluated at new query points. CORAL~\citep{serrano2023coral}, for example, learns a map between latent codes for the input and output fields and decodes the output at the requested coordinates.

\paragraph{Our map.}
Our map belongs to the first family. It is one network, shared across the dataset, and a new shape costs a forward pass.
Two things separate it from Geo-FNO, which has also been used with a spherical latent domain~\citep{li2023gino}.
First, Geo-FNO learns a displacement at each point. Our map is the flow of a velocity field, so the continuous deformation is invertible by construction. A penalty keeps its endpoints close to the sphere before projection.
Second, the points are resampled onto the latent grid before the operator is applied, which allows us to use an FFT.
On an unstructured mesh, Geo-FNO leaves the deformed points scattered, so its input transform is a direct non-uniform DFT evaluated as a dense product.
For $N$ points and $M$ retained modes, this costs $O(NM)$. That transform is linear in $N$ for fixed $M$, and every additional mode costs another $O(N)$.
An FFT returns the whole grid spectrum at once. Resampling is what makes it applicable. Section~\ref{sec:analysis} measures the reconstruction error introduced by this step.

\section{Method}
\label{sec:method}

\subsection{Formulation}

The input consists of the coordinates and normals of surface points, whose number and locations vary from car to car.
From this geometry, we predict either the pressure $p(x_i)$ at each queried surface point $x_i$ or a single drag coefficient $C_D$ for the entire car.

\begin{figure}[htbp]
  \centering
  \includegraphics[width=\columnwidth]{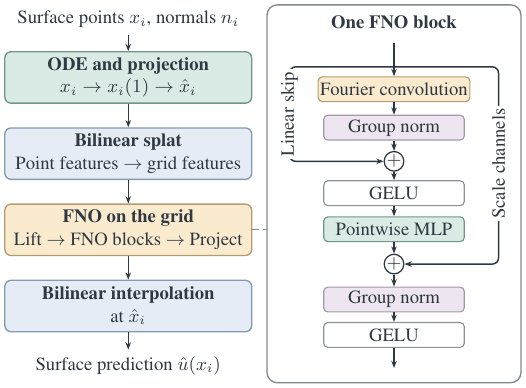}
  \caption{Surface field prediction with a learned map to the sphere. Splatting and interpolation use the same projected locations $\hat{x}_i$. The final FNO block omits the two GELU steps shown here.}
  \label{fig:field-pipeline}
\end{figure}

We present a Learned Continuous–Flow Reparametrisation method.
We learn a map from the object's surface to the latent sphere. Using this map, we splat surface features onto a regular spherical grid with $\mathcal S_\theta$, apply a Fourier neural operator $\mathcal K$ on the grid, and interpolate the predicted field back to the original surface points with $\mathcal I_\theta$. The resulting surface operator is
\[
  \mathcal A = \mathcal I_\theta \circ \mathcal K \circ \mathcal S_\theta.
\]
Figure~\ref{fig:field-pipeline} illustrates this composition for surface field prediction.

\paragraph{Transfer to the latent grid.}
We denote the $N$ surface points and their associated field values by
\[
  \mathcal{P} = \{(x_i, u_i)\}_{i=1}^{N},
  \qquad u_i = u(x_i),
\]
where $x_i \in \mathbb{R}^3$ is the position of the $i$th point and $u$ is the surface field. For pressure prediction, $u=p$.

We discretise the latent sphere $\mathbb{S}^2$ using an equiangular latitude--longitude grid
\[
  \mathcal{G} = \{g_{jk} \mid 1 \le j \le H,\ 1 \le k \le W\}
  \subset \mathbb{S}^2 \subset \mathbb{R}^3,
\]
where $g_{jk}$ is the point at the $j$th latitude and $k$th longitude.

We transfer the field from the object's surface to the latent space in two successive steps.
\begin{enumerate}
\item For each surface point $x_i$, we integrate the following ODE from $t=0$ to $t=1$:
\[
  \frac{\mathrm{d}x_i(t)}{\mathrm{d}t} = \phi_\theta(x_i(t), t),
  \qquad x_i(0) = x_i,
\]
where $\phi_\theta$ is a neural network representing the velocity field \citep{chen2018neuralode}.
Its inputs are the current coordinates $x_i(t)$ of the point and the integration time $t$.
The endpoint defines the map $T(x_i) = x_i(1)$.
\item We project each trajectory endpoint $x_i(1)$ onto the unit sphere by setting
\[
  \hat{x}_i = \frac{x_i(1)}{\lVert x_i(1)\rVert_2} \in \mathbb{S}^2.
\]
We then use bilinear splatting to distribute the field value $u_i$ at each projected point $\hat{x}_i$ to the four neighbouring nodes of the grid $\mathcal{G}$.
Let $a_i$ and $b_i$ be the fractional parts of its row and column coordinates. The four corner weights, in row and column order, are
\[
  \begin{pmatrix}
    (1-a_i)(1-b_i) & (1-a_i)b_i \\
    a_i(1-b_i)     & a_i b_i
  \end{pmatrix}.
\]
For point $i$, we assign each matrix entry to the grid node at the corresponding corner and denote the sum of the entries assigned to $g_{jk}$ by $\omega_{i,jk}$ (zero if there are none). At each node with positive total weight, the field value is the weighted mean
\[
  u_{\mathcal{G}}(g_{jk}) =
  \frac{\sum_{i=1}^{N} \omega_{i,jk}\,u_i}
       {\sum_{i=1}^{N} \omega_{i,jk}}.
\]
\end{enumerate}

Let $n_i$ denote the surface normal at $x_i$.
We apply the same splatting rule componentwise to the geometric features $f_i=[x_i,\,x_i(1),\,n_i]\in\mathbb{R}^9$.

\paragraph{Field prediction on the latent grid.}
We feed the splatted features $f_{\mathcal G}$ to a two-dimensional Fourier neural operator (FNO)~\citep{li2021fno} with $L$ blocks. At each grid node $g$, a learned map $P$ combines the features with the normalised grid coordinates $c(g)$ to give $h_0(g)=P([f_{\mathcal G}(g),c(g)])$.
Each FNO block applies a Fourier convolution along the two axes of the latitude--longitude grid,
\[
  K_\ell h = \mathcal F^{-1}\!\left(R_\ell\,\mathcal F h\right)+b_\ell,
\]
where $\mathcal F$ is the two-dimensional discrete Fourier transform, $R_\ell$ is a learned channel-mixing matrix at each retained mode and is zero elsewhere, and $b_\ell$ is a learned bias. Each block also includes a pointwise MLP, skip connections and group normalisation.
After $L$ blocks, we apply a learned map $Q$ at each grid node $g$ to obtain the output field $\hat u_{\mathcal G}(g)=Q(h_L(g))$.

\paragraph{Transfer back to the surface.}
We then use bilinear interpolation to evaluate the predicted field on the grid $\mathcal G$ at each projected ODE endpoint $\hat{x}_i$. Using the same four corner weights as in splatting, we obtain
\[
  \hat u(\hat{x}_i) = \sum_{j=1}^{H}\sum_{k=1}^{W}
  \omega_{i,jk}\,\hat u_{\mathcal G}(g_{jk}).
\]
When we transfer the field to the latent space, we retain the correspondence between each surface point $x_i$ and its image $\hat{x}_i$.
Through this correspondence, the values $\hat u(\hat{x}_i)$ give the predicted field at the points $x_i$ on the object.
The input and output points can also be sampled independently on the same surface.
Once the FNO has predicted $\hat u_{\mathcal G}$ from a sparse input, we can evaluate pressure at any additional surface point $q$ by mapping it to $T(q)/\lVert T(q)\rVert_2$ and interpolating the same grid field there.

\subsection{Training procedure}

To predict surface pressure $p$, the operator takes the splatted geometric features $f_{\mathcal G}$ as input, and we interpolate its output back to each surface point.
To predict the drag coefficient $C_D$, following OTNO~\citep{li2025otno}, we first take the dot product of each surface normal $n_i$ with the unit flow direction $d$,
\[
  w_i = n_i \cdot d, \qquad d=(1,0,0).
\]
We splat these signed weights onto the latent grid and multiply the features at each node $g$ by the corresponding weight, giving the operator input $w_{\mathcal G}(g)\,f_{\mathcal G}(g)$ with $w_{\mathcal G}=\mathcal S_\theta(w)$.
We interpolate the operator output back to the surface and average over all $N$ points to predict $C_D$.

For both tasks, we first train the surface-to-sphere mapping on its own and then jointly optimise the mapping and the operator.

\paragraph{Mapping pretraining.}
For both pressure and drag, we pretrain the velocity field $\phi_\theta$ using flow matching~\citep{lipman2023flowmatching} on point correspondences obtained by optimal transport.
For each training geometry, we precompute a correspondence between its surface points and the spherical grid by solving a discrete optimal transport problem.
For each surface point $x_i$, we use the corresponding grid point $y_i$ as the endpoint of a straight-line path for flow matching.
\[
  z_i(t) = (1-t)x_i + t y_i, \qquad t \in [0,1].
\]
For a batch of $B$ point pairs, we independently sample $t_i$ uniformly from $[0,1]$ and train the velocity field by minimising the mean squared error against the path velocity $y_i-x_i$,
\[
  \mathcal{L}_{\mathrm{FM}}(\theta)
  = \frac{1}{3B}\sum_{i=1}^{B}
    \bigl\lVert \phi_\theta(z_i(t_i),t_i) - (y_i-x_i) \bigr\rVert_2^2.
\]
At inference, we obtain the map for a new geometry by integrating the learned velocity field, without solving a new optimal transport problem.

\paragraph{Joint training.}
We initialise the surface-to-sphere map with the velocity field pretrained in the first stage and jointly optimise the mapping and the FNO end to end.
We minimise a weighted sum of three losses. For each geometry, we compute the data loss on standardised predictions and targets:
\[
  \mathcal L_{\mathrm{data}} =
  \begin{cases}
    \dfrac{\lVert \hat u-u\rVert_2}{\lVert u\rVert_2},
      & \text{for a surface field } u,\\[5pt]
    (\hat C_D-C_D)^2, & \text{for the drag coefficient } C_D.
  \end{cases}
\]
The sphere loss penalises the radial distance of each ODE endpoint $x_i(1)$ from the unit sphere before projection,
\[
  \mathcal L_{\mathrm{sph}}
  = \frac{1}{N}\sum_{i=1}^{N}\bigl(\lVert x_i(1)\rVert_2-1\bigr)^2.
\]
The uniformity loss discourages the projected points $\hat x_i$ from concentrating in one region of the sphere~\citep{wang2020hypersphere},
\[
  \mathcal L_{\mathrm{unif}}
  = \log \mathbb E_{i\ne j}
    \bigl[\exp\bigl(-2\lVert\hat x_i-\hat x_j\rVert_2^2\bigr)\bigr],
\]
where the expectation averages over distinct pairs in a random subset of the surface points. The joint training loss is
\[
  \mathcal L_{\mathrm{joint}} =
  \mathcal L_{\mathrm{data}}
  + w_{\mathrm{sph}}\mathcal L_{\mathrm{sph}}
  + w_{\mathrm{unif}}\mathcal L_{\mathrm{unif}}.
\]
Gradients from the prediction loss update both the operator and the velocity field $\phi_\theta$, allowing the pretrained mapping to adapt to the field prediction task.
We use a lower learning rate for the velocity field than for the operator.

\section{Experiments}
\label{sec:exp}

We use a single NVIDIA A100 GPU for each training run of our model.

\paragraph{Datasets.}
We evaluate our model on ShapeNet-Car~\citep{umetani2018learning}, AhmedML~\citep{ashton2024ahmedml}, and DrivAerNet++~\citep{drivaernetpp}.
We predict surface pressure $p$ on all three datasets and additionally predict the drag coefficient $C_D$ on DrivAerNet++.
We apply voxel downsampling to the AhmedML and DrivAerNet++ surfaces and retain the original ShapeNet-Car mesh vertices.
For each dataset, we use the same split across methods and training seeds.
Following the ShapeNet-Car protocol of GINO~\citep{li2023gino} and OTNO~\citep{li2025otno}, all methods train on 500 cars and use the same 111-car split for epoch selection and reporting.
For DrivAerNet++, we follow the official train/validation/test split.
For AhmedML, we fix three disjoint sets of geometries.
Table~\ref{tab:datasets} lists the number of surface points per geometry and the split sizes.

\begin{table}[htbp]
\centering
\small
\setlength{\tabcolsep}{3pt}
\caption{Surface points per geometry after preprocessing (mean $\pm$ standard deviation) and number of geometries per split.}
\label{tab:datasets}
\begin{tabular}{@{}lrrrr@{}}
\toprule
Dataset & Points & Train & Val. & Test \\
\midrule
ShapeNet-Car & $3{,}586$ & 500 & 111 & --- \\
AhmedML & $135\mathrm{k}\pm21\mathrm{k}$ & 400 & 50 & 50 \\
DrivAerNet++ & $16.2\mathrm{k}\pm1.7\mathrm{k}$ & 5,817 & 1,148 & 1,154 \\
\bottomrule
\end{tabular}
\end{table}

\paragraph{Baselines.}
We compare our method with OTNO~\citep{li2025otno}, GINO~\citep{li2023gino}, and RegDGCNN for drag~\citep{drivaernet} and surface pressure~\citep{drivaernetpp}.
OTNO is the closest baseline because both methods map the surface to a regular two-dimensional grid and apply a Fourier neural operator on that grid.
For surface pressure prediction, we compare our FNO variant with OTNO using the same FNO hyperparameters.

\subsection{Results}
\label{sec:results}

\paragraph{Surface pressure prediction.}
Table~\ref{tab:pressure} reports the relative $L_2$ error in physical pressure, computed for each geometry and averaged over the evaluation set.
In our runs, the learned map yields a lower error than OTNO on ShapeNet-Car and AhmedML.
On DrivAerNet++, OTNO has the lowest error ($17.00\%$), followed by our method ($18.06\%$).
Our method and OTNO both achieve lower relative $L_2$ errors than GINO and RegDGCNN on all three pressure benchmarks.
RegDGCNN uses a subset of 10,000 surface points per geometry on AhmedML and 5,000 on DrivAerNet++.

\begin{table}[htbp]
\centering
\small
\setlength{\tabcolsep}{3pt}
\caption{Surface pressure error (relative $L_2$ in \%, lower is better). Our FMSNO and OTNO runs give the mean and sample standard deviation over three seeds. The published OTNO result is from \citet{li2025otno}. For ShapeNet-Car, results are reported on the 111-geometry benchmark split used by GINO and OTNO. All methods use the same 50 test geometries on AhmedML.}
\label{tab:pressure}
\begin{tabular}{@{}lccc@{}}
\toprule
Method & ShapeNet-Car & AhmedML & DrivAerNet++ \\
\midrule
Ours    & $\mathbf{6.73}\pm0.15$ & $\mathbf{3.83}\pm0.13$ & $18.06\pm0.15$ \\
OTNO (our runs) & $7.16\pm0.04$  & $\mathbf{4.01}\pm0.12$ & $\mathbf{17.00}\pm0.06$ \\
OTNO (paper) & $\mathbf{6.70}$ & --- & --- \\
GINO    & $7.71$                 & $5.21$                               & $19.33$ \\
RegDGCNN & $9.15$                & $6.76$                               & $20.86$ \\
\bottomrule
\end{tabular}
\end{table}

\paragraph{Drag coefficient prediction.}
On the official DrivAerNet++ test set, FMSNO achieves an $R^2$ of $0.883$ with FNO and $0.906$ with SFNO, compared with $0.887$ for OTNO, $0.874$ for GINO and $0.665$ for RegDGCNN (Table~\ref{tab:drag}).

\begin{table}[htbp]
\centering
\small
\caption{Drag coefficient prediction on DrivAerNet++. Both FMSNO variants use FM pretraining. FMSNO and OTNO results use seed 42.}
\label{tab:drag}
\begin{tabular}{@{}lcc@{}}
\toprule
Method & $R^2$~$\uparrow$ & MSE ($\times 10^{-5}$)~$\downarrow$ \\
\midrule
Ours (FNO)  & $0.883$ & $4.84$ \\
Ours (SFNO) & $\mathbf{0.906}$ & $\mathbf{3.91}$ \\
OTNO (FNO)  & $0.887$ & $4.69$ \\
GINO        & $0.874$                  & $5.22$ \\
RegDGCNN    & $0.665$                  & $13.9$ \\
\bottomrule
\end{tabular}
\end{table}

\subsection{Inference scalability}
\label{sec:scaling}

\begin{figure*}[t]
\centering
\includegraphics[width=0.92\textwidth]{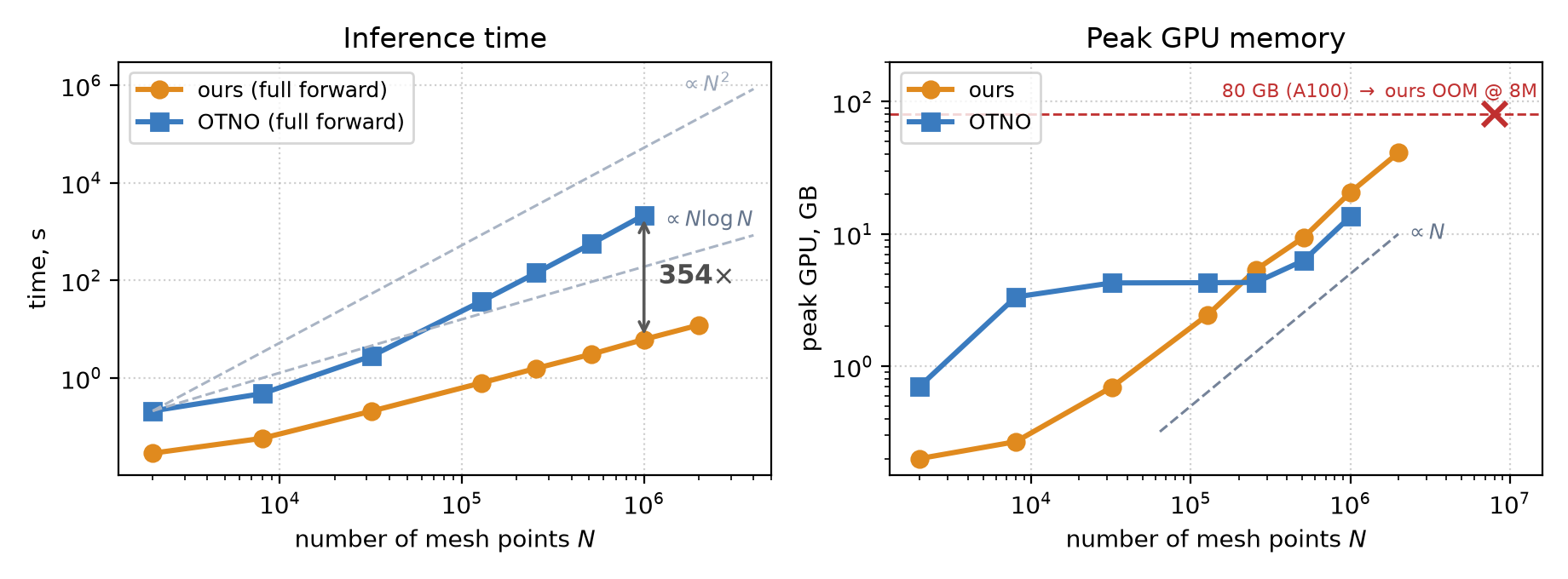}
\caption{End-to-end inference time (left) and peak GPU memory (right) on one A100 (80\,GB). Both methods use the same FNO grid size at each resolution. The grey dashed line on the right indicates the linear memory scaling of both implementations. The two smallest meshes are real, and meshes with $N\geq 32{,}000$ are synthetic. At one million surface points, our method is $354\times$ faster than OTNO, including its transport solve for the new geometry.}
\label{fig:inference_scaling}
\end{figure*}

On a new synthetic surface with one million points, our method achieves a $354\times$ end-to-end inference speedup over OTNO (Fig.~\ref{fig:inference_scaling}).
The timing includes the optimal transport solve that OTNO performs for the new geometry.

\paragraph{Complexity.}
For each new geometry with $N$ surface points, OTNO~\citep{li2025otno} solves a new optimal transport problem at inference.
Each Sinkhorn iteration evaluates pairwise costs between the $N$ surface points and a latent grid with $\Theta(N)$ nodes, giving $O(N^2)$ time for a fixed number of iterations.
Our learned map takes $O(N)$ time because it integrates a fixed number of ODE steps independently for each point.
Bilinear splatting and interpolation also take $O(N)$ time, since each point interacts with four grid nodes.
With the FNO backbone, the FFTs therefore determine the overall inference complexity of $O(N\log N)$.
Figure~\ref{fig:inference_scaling} shows how inference time and peak GPU memory vary with $N$.

\section{Method analysis and ablations}
\label{sec:analysis}
\label{sec:ablations}

\paragraph{How much information does splatting lose?}
Bilinear splatting averages the pressures of surface points that contribute to the same grid node.
This averaging can smooth local pressure differences. In particular, if two surface points map to the same location on the latent grid, bilinear interpolation returns the same prediction at both, even when their true pressures differ.
For each geometry, we splat the true surface pressure values $p$ onto the grid and interpolate them directly back to the original surface points to obtain the reconstructed values $p_{\mathrm{rec}}$.
The relative resampling error for that geometry is
\[
  e_{\mathrm{res}} = \frac{\lVert p_{\mathrm{rec}}-p\rVert_2}{\lVert p\rVert_2}.
\]
We average $e_{\mathrm{res}}$ over the test geometries and report the result as a percentage.
Table~\ref{tab:fm-resampling} shows that, with the default uniformity weight $w_{\mathrm{unif}}=1$, joint training reduces this error from $19.4\%$ to $10.2\%$ on DrivAerNet++.
Over the same test geometries, the fraction of surface points sharing their nearest grid node with another point rises from $23.2\%$ to $36.8\%$.
This suggests that the map learns to use the available grid resolution more effectively to represent the pressure field.

\begin{table}[htbp]
\centering
\small
\caption{Effect of the uniformity weight on 1,154 DrivAerNet++ test cars (percentages, errors in relative $L_2$). The baseline prediction error (weight 1) is the mean over seeds 1, 2 and 42. All other entries use seed 42 and the same initial FM map.}
\label{tab:fm-resampling}
\setlength{\tabcolsep}{3pt}
\begin{tabular}{@{}cccccc@{}}
\toprule
 & \multicolumn{2}{c}{Resampling} & \multicolumn{2}{c}{Collisions} & Prediction \\
\cmidrule(lr){2-3}\cmidrule(lr){4-5}
$w_{\mathrm{unif}}$ & Before & After & Before & After & Rel.\ $L_2$ \\
\midrule
1 & 19.4 & \textbf{10.2} & 23.2 & 36.8 & \textbf{18.06} \\
2 & 19.4 & 10.3 & 23.2 & 35.9 & 18.09 \\
3 & 19.4 & 10.9 & 23.2 & 37.2 & 18.39 \\
\bottomrule
\end{tabular}
\end{table}

\begin{figure}[htbp]
\centering
\includegraphics[width=\columnwidth]{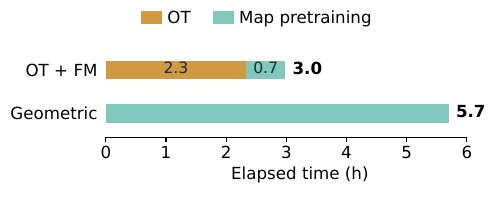}
\caption{DrivAerNet++ map preparation time on one GPU. OT preprocessing covers all 8,119 geometries. Only pairs from the 5,817 training geometries contribute to FM weight updates.}
\label{fig:training-time}
\end{figure}

\paragraph{Is optimal transport needed for map pretraining?}
We replace flow matching on OT correspondences with a geometric warm-up that uses only the sphere and uniformity losses,
\[
  \mathcal L_{\mathrm{geom}}(\theta)
  = \mathcal L_{\mathrm{sph}}(\theta) + \mathcal L_{\mathrm{unif}}(\theta).
\]
On DrivAerNet++, we pretrain the map for 50 epochs with this objective or for 200 epochs with FM, then use the same 100-epoch joint training procedure in both runs.
Before joint training, the geometric map has nearly twice the resampling error of the FM map ($38.8\%$ versus $19.4\%$).
Joint training narrows the gap to $12.5\%$ versus $10.2\%$. Table~\ref{tab:map-init} compares final pressure prediction errors across the three datasets. FM gives a lower error on all three.
Preparing the DrivAerNet++ map took 2~h 59~min with FM, including the one-time OT solve, and 5~h 42~min with the geometric warm-up (Fig.~\ref{fig:training-time}).

\begin{table}[htbp]
\centering
\small
\caption{Test pressure error after joint training (relative $L_2$ in \%). FM results are means over seeds 1, 2 and 42. Geometric warm-up uses seed 42.}
\label{tab:map-init}
\setlength{\tabcolsep}{3pt}
\begin{tabular}{@{}lcc@{}}
\toprule
Dataset & FM & Geometric warm-up \\
\midrule
ShapeNet-Car & $6.73$ & $7.09$ \\
AhmedML      & $3.83$ & $4.06$ \\
DrivAerNet++ & $18.06$ & $19.07$ \\
\bottomrule
\end{tabular}
\end{table}

\begin{figure}[htbp]
\centering
\includegraphics[width=\columnwidth]{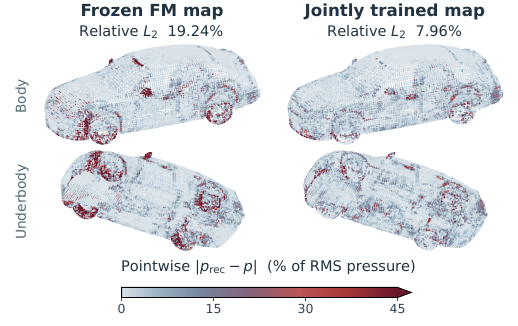}
\caption{Reconstruction of the reference pressure on the DrivAerNet++ test car in Fig.~\ref{fig:drivaer-pipeline}, chosen for its near-median map deformation. Both maps resample the same 16,174 points. Colours show absolute reconstruction error on a shared scale, as a percentage of the root mean square reference pressure on this car.}
\label{fig:pressure-resampling}
\end{figure}

\paragraph{Does joint training adapt the map to the pressure field?}
Both variants train the FNO for 100 epochs and share the same initial FM map at seed 42. Joint training better preserves pressure under splatting and interpolation and lowers the reported prediction error by $6.46$ percentage points (Table~\ref{tab:joint-map}).
On the test geometry in Fig.~\ref{fig:pressure-resampling}, joint training reduces reconstruction error around the wheels. The lowest third of the surface points by height accounts for $52\%$ of the reduction in total squared reconstruction error on this geometry.

\begin{table}[htbp]
\centering
\small
\caption{Pressure errors on 1,154 DrivAerNet++ test cars (mean relative $L_2$ in \%). Joint prediction error is the mean over seeds 1, 2 and 42. Resampling and frozen-map results use seed 42.}
\label{tab:joint-map}
\begin{tabular}{@{}lcc@{}}
\toprule
 & \multicolumn{2}{c}{Mean relative $L_2$ error (\%) $\downarrow$} \\
\cmidrule(l){2-3}
Training & Resampling & Prediction \\
\midrule
FNO only (frozen map) & 19.4 & 24.52 \\
FNO and map jointly & \textbf{10.2} & \textbf{18.06} \\
\bottomrule
\end{tabular}
\end{table}

\begin{figure}[htbp]
\centering
\includegraphics[width=\columnwidth]{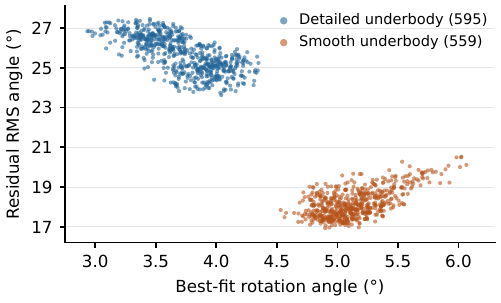}
\caption{Map displacement from FM pretraining to joint training on DrivAerNet++. Each point shows one test car's fitted rotation angle and the RMS angular displacement remaining after alignment. Colours indicate the underbody type from the dataset labels. The smooth group includes 183 fastbacks.}
\label{fig:map-motion-clusters}
\end{figure}

\paragraph{How does joint training move points on the sphere?}
For each of the 1,154 DrivAerNet++ test cars, we compare where the same surface vertices map on the sphere after FM pretraining and after joint training.
If joint training simply rotates the entire set of points, a single rotation should bring every point back to its pretrained position.
We fit one rotation per car by rotating all jointly trained points together to minimise the mean squared angular distance to their respective pretrained positions.
On average, this alignment removes only $3.27\%$ of each car's mean squared angular displacement. Joint training therefore mainly changes the positions of points relative to one another.
Across cars, the fitted angle averages $4.41^\circ$. The root mean square angular displacement averages $22.39^\circ$ before alignment and $22.05^\circ$ afterwards.
The test cars form two distinct clusters in Fig.~\ref{fig:map-motion-clusters}. The fastbacks with a detailed underbody undergo smaller rotations and larger residual displacements than the cars in the seven smooth-underbody configurations.
That is a clear sign that joint training influences the mapping in an object-dependent way.

\begin{figure}[htbp]
\centering
\includegraphics[width=\columnwidth]{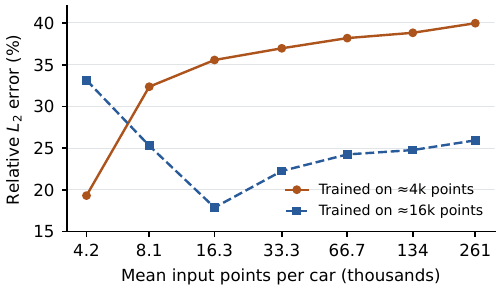}
\caption{Surface pressure error on DrivAerNet++ for two FMSNO models trained on approximately 4,000 and 16,000 points per car (seed 42). At each of seven voxel resolutions, we compute the relative $L_2$ error over all retained points on each surface and report the mean over the same 1,154 test cars.}
\label{fig:coarse-fine-pressure}
\end{figure}

\paragraph{How well does a model trained on coarse meshes predict pressure on finer meshes?}
We evaluated two FMSNO models on the same 1,154 DrivAerNet++ test cars at seven voxel resolutions (Fig.~\ref{fig:coarse-fine-pressure}).
For the model trained on roughly 4,000 points per car, the mean relative $L_2$ pressure error more than doubled, from $19.32\%$ at its training resolution to $39.97\%$ on the finest test meshes (about 261,000 points per car).
The model trained on roughly 16,000 points lost less accuracy as the meshes became finer. Its error rose from $17.89\%$ at its training resolution to $25.93\%$ at the finest resolution.

\paragraph{How well does the model generalise to unseen body styles?}
We held out each of the two least frequent DrivAerNet++ body styles, Estateback and Notchback, in turn and trained our model on the other two styles.
For each comparison, we trained a control on all three styles, keeping all training cars of the target style and subsampling the other styles to match the total number of training cars.
We selected both checkpoints on the same validation cars from the two remaining styles and evaluated them on the same test cars of the target style (Table~\ref{tab:body-style-transfer}).
Holding out the target style increased pressure error by 4.90 percentage points for Estateback and 2.34 for Notchback. On test cars of the two remaining styles, the error changed by at most 0.12 percentage points.

\begin{table}[htbp]
\centering
\small
\caption{Pressure prediction on the target body style in DrivAerNet++ (seed 42). Train gives the number of training cars per model. Errors are mean relative $L_2$ over the target test cars, with the target style included in or held out of training.}
\label{tab:body-style-transfer}
\setlength{\tabcolsep}{4pt}
\begin{tabular}{@{}lrrcc@{}}
\toprule
 & \multicolumn{2}{c}{Cars} & \multicolumn{2}{c}{Relative $L_2$ (\%) $\downarrow$} \\
\cmidrule(lr){2-3}\cmidrule(l){4-5}
Target style & Train & Test & Included & Held out \\
\midrule
Estateback & 4,856 & 210 & 19.93 & 24.83 \\
Notchback  & 4,743 & 166 & 18.56 & 20.90 \\
\bottomrule
\end{tabular}
\end{table}

\paragraph{How does the latent grid resolution affect pressure prediction?}
For a geometry with $N$ surface points, we set the grid dimensions to $H=W=\lceil\sqrt{\alpha N}\rceil$ and vary the expansion factor $\alpha$.

\begin{table}[htbp]
\centering
\small
\caption{Effect of grid resolution on pressure error over 1,154 DrivAerNet++ test cars. The baseline ($\alpha=3$) is the mean over seeds 1, 2 and 42. Finer grids use seed 42.}
\label{tab:latent-grid-resolution}
\setlength{\tabcolsep}{4pt}
\begin{tabular}{@{}lcccc@{}}
\toprule
Expansion factor $\alpha$ & 3 (baseline) & 4 & 5 & 6 \\
\midrule
Relative $L_2$ (\%) $\downarrow$ & 18.06 & 18.21 & \textbf{17.46} & 18.07 \\
\bottomrule
\end{tabular}
\end{table}

Pressure errors are similar across the tested resolutions and show no consistent trend as the grid is refined (Table~\ref{tab:latent-grid-resolution}).
One possible explanation is that a finer grid can separate nearby surface points that would otherwise be averaged together, but it also spreads the same input over more grid nodes.
Each point contributes to at most four nodes, so at $\alpha=6$ at least one third of the nodes receive no surface features from the splat.
Our choice of $\alpha=3$ by default is motivated by GPU memory and computational time budget limitations and overall good quality/computational cost balance for this setting.

\paragraph{Does the choice of operator affect prediction quality?}
FNO gives lower pressure error, while SFNO gives a higher $R^2$ for drag (Table~\ref{tab:operator-choice}).

\begin{table}[htbp]
\centering
\small
\caption{FM-pretrained operators on DrivAerNet++. Map LR is relative to operator LR. FNO pressure error is the mean over seeds 1, 2 and 42. Other entries use seed 42.}
\label{tab:operator-choice}
\setlength{\tabcolsep}{4pt}
\begin{tabular}{@{}ccccc@{}}
\toprule
Task & Map LR scale & Metric & FNO & SFNO \\
\midrule
Pressure & 0.10 & Rel.\ $L_2$ (\%) $\downarrow$ & \textbf{18.06} & 18.20 \\
$C_D$ (FM) & 0.03 & $R^2$ $\uparrow$ & 0.8832 & \textbf{0.9058} \\
\bottomrule
\end{tabular}
\end{table}

\section{Limitations}
\label{sec:limits}

\paragraph{Loss of detail during resampling.}
Around mirrors, wheel arches and spoilers, surface regions
on opposite sides of a narrow gap can be close in space yet carry different pressures.
After the flow and projection onto the sphere, points from these regions can contribute to the same
grid nodes. Bilinear splatting averages their geometric features at each shared node before the operator sees them.
In \S\ref{sec:analysis}, we apply the same splatting and interpolation to the reference pressure.
Table~\ref{tab:fm-resampling} reports the relative resampling error, and Fig.~\ref{fig:pressure-resampling}
shows where pressure detail is lost on one test car.
This smoothing may help explain why OTNO achieves lower pressure error on DrivAerNet++
(Table~\ref{tab:pressure}).

\paragraph{Shape-independent map.}
The velocity field $\phi_\theta(x,t)$ depends only on position and time,
so points with the same initial coordinates on different cars follow the same trajectory even when the surrounding surfaces differ.

\section{Conclusion}
We replaced the per-shape optimal transport map of surface neural operators with a continuous
normalising flow that is learned once and applied by a forward pass, so that the map for a new mesh
costs $O(N)$ instead of an $O(N^2)$ solve and full inference takes $O(N\log N)$ time
with the FNO backbone shared by both pipelines. In our runs, FMSNO achieves lower pressure error than OTNO on two of the three
datasets. At $10^6$ surface points inference is $354\times$ faster.
On DrivAerNet++, we measure how accurately the reference pressure can be reconstructed by splatting
it onto the latent grid and interpolating it back to the surface. Joint training reduces the mean
relative $L_2$ error of this reconstruction from $19.4\%$ to $10.2\%$ (Table~\ref{tab:fm-resampling}).
With a shape-conditioned velocity field, points at the same initial coordinates could follow different trajectories on different cars.

\section*{LLM Usage Statement}
We used OpenAI ChatGPT for editing, refining, and polishing the manuscript
and for editing and polishing the code implementation. ChatGPT also
provided reviewer-style feedback on the manuscript.
We did not use AI tools to develop research ideas or hypotheses, formulate
claims, or draw conclusions. We take full responsibility for all submitted content.

{
    \small
    \bibliographystyle{ieeenat_fullname}
    \bibliography{main}

@String(CVPR= {IEEE Conf. Comput. Vis. Pattern Recog.})

@String(ICLR = {Int. Conf. Learn. Represent.})

@String(CVPR  = {CVPR})

@String(ICLR  = {ICLR})

@inproceedings{chen2018neuralode,
  title     = {Neural Ordinary Differential Equations},
  author    = {Chen, Ricky T. Q. and Rubanova, Yulia and Bettencourt, Jesse and Duvenaud, David},
  booktitle = {Advances in Neural Information Processing Systems (NeurIPS)},
  volume    = {31},
  year      = {2018},
  note      = {arXiv:1806.07366}
}

@inproceedings{lipman2023flowmatching,
  title     = {\href{https://arxiv.org/abs/2210.02747}{Flow Matching for Generative Modeling}},
  author    = {Lipman, Yaron and Chen, Ricky T. Q. and Ben-Hamu, Heli and Nickel, Maximilian and Le, Matt},
  booktitle = {International Conference on Learning Representations (ICLR)},
  year      = {2023}
}

@inproceedings{bonev2023sfno,
  title     = {\href{https://proceedings.mlr.press/v202/bonev23a.html}{Spherical {Fourier} Neural Operators: Learning Stable Dynamics on the Sphere}},
  author    = {Bonev, Boris and Kurth, Thorsten and Hundt, Christian and Pathak, Jaideep and Baust, Maximilian and Kashinath, Karthik and Anandkumar, Anima},
  booktitle = {Proceedings of the 40th International Conference on Machine Learning},
  volume    = {202},
  series    = {Proceedings of Machine Learning Research},
  pages     = {2806--2823},
  year      = {2023}
}

@inproceedings{li2021fno,
  title     = {{Fourier} Neural Operator for Parametric Partial Differential Equations},
  author    = {Li, Zongyi and Kovachki, Nikola and Azizzadenesheli, Kamyar and Liu, Burigede and Bhattacharya, Kaushik and Stuart, Andrew and Anandkumar, Anima},
  booktitle = {International Conference on Learning Representations (ICLR)},
  year      = {2021}
}

@inproceedings{wang2020hypersphere,
  title     = {Understanding Contrastive Representation Learning through Alignment and Uniformity on the Hypersphere},
  author    = {Wang, Tongzhou and Isola, Phillip},
  booktitle = {International Conference on Machine Learning (ICML)},
  pages     = {9929--9939},
  year      = {2020}
}

@article{umetani2018learning,
  title   = {\href{https://doi.org/10.1145/3197517.3201325}{Learning Three-Dimensional Flow for Interactive Aerodynamic Design}},
  author  = {Umetani, Nobuyuki and Bickel, Bernd},
  journal = {ACM Transactions on Graphics},
  volume  = {37},
  number  = {4},
  pages   = {89:1--89:10},
  year    = {2018}
}

@article{ashton2024ahmedml,
  title   = {\href{https://arxiv.org/abs/2407.20801}{{AhmedML}: High-Fidelity Computational Fluid Dynamics Dataset for Incompressible, Low-Speed Bluff Body Aerodynamics}},
  author  = {Ashton, Neil and Maddix, Danielle C. and Gundry, Samuel and Shabestari, Parisa M.},
  journal = {arXiv preprint arXiv:2407.20801},
  year    = {2024}
}

@inproceedings{drivaernetpp,
  title     = {\href{https://arxiv.org/abs/2406.09624}{{DrivAerNet++}: A Large-Scale Multimodal Car Dataset with Computational Fluid Dynamics Simulations and Deep Learning Benchmarks}},
  author    = {Elrefaie, Mohamed and Morar, Florin and Dai, Angela and Ahmed, Faez},
  booktitle = {Advances in Neural Information Processing Systems},
  volume    = {37},
  year      = {2024}
}

@article{li2025otno,
  title   = {\href{https://www.jmlr.org/papers/v26/25-1380.html}{Geometric Operator Learning with Optimal Transport}},
  author  = {Li, Xinyi and Li, Zongyi and Kovachki, Nikola and Anandkumar, Anima},
  journal = {Journal of Machine Learning Research},
  volume  = {26},
  number  = {282},
  pages   = {1--38},
  year    = {2025}
}

@inproceedings{li2023gino,
  title     = {\href{https://arxiv.org/abs/2309.00583}{Geometry-Informed Neural Operator for Large-Scale {3D} {PDEs}}},
  author    = {Li, Zongyi and Kovachki, Nikola Borislavov and Choy, Chris and Li, Boyi and Kossaifi, Jean and Otta, Shourya Prakash and Nabian, Mohammad Amin and Stadler, Maximilian and Hundt, Christian and Azizzadenesheli, Kamyar and Anandkumar, Anima},
  booktitle = {Advances in Neural Information Processing Systems},
  year      = {2023}
}

@article{drivaernet,
  title   = {\href{https://arxiv.org/abs/2403.08055}{{DrivAerNet}: A Parametric Car Dataset for Data-Driven Aerodynamic Design and Prediction}},
  author  = {Elrefaie, Mohamed and Dai, Angela and Ahmed, Faez},
  journal = {arXiv preprint arXiv:2403.08055},
  year    = {2024}
}

@techreport{slotnick2014cfdvision,
  title       = {{CFD} Vision 2030 Study: A Path to Revolutionary Computational Aerosciences},
  author      = {Slotnick, Jeffrey P. and Khodadoust, Abdollah and Alonso, Juan and Darmofal, David and Gropp, William and Lurie, Elizabeth and Mavriplis, Dimitri J.},
  institution = {NASA},
  number      = {NASA/CR-2014-218178},
  year        = {2014}
}

@article{azizzadenesheli2024neuralop,
  title   = {Neural operators for accelerating scientific simulations and design},
  author  = {Azizzadenesheli, Kamyar and Kovachki, Nikola and Li, Zongyi and Liu-Schiaffini, Miguel and Kossaifi, Jean and Anandkumar, Anima},
  journal = {Nature Reviews Physics},
  volume  = {6},
  pages   = {320--328},
  year    = {2024},
  doi     = {10.1038/s42254-024-00712-5}
}

@article{li2023geofno,
  title   = {Fourier Neural Operator with Learned Deformations for {PDEs} on General Geometries},
  author  = {Li, Zongyi and Huang, Daniel Zhengyu and Liu, Burigede and Anandkumar, Anima},
  journal = {Journal of Machine Learning Research},
  volume  = {24},
  number  = {388},
  pages   = {1--26},
  year    = {2023},
  note    = {arXiv:2207.05209}
}

@article{komatitsch1999sem,
  title   = {Introduction to the spectral element method for three-dimensional seismic wave propagation},
  author  = {Komatitsch, Dimitri and Tromp, Jeroen},
  journal = {Geophysical Journal International},
  volume  = {139},
  number  = {3},
  pages   = {806--822},
  year    = {1999},
  doi     = {10.1046/j.1365-246x.1999.00967.x}
}

@book{boyd2001spectral,
  title     = {Chebyshev and {Fourier} Spectral Methods},
  author    = {Boyd, John P.},
  edition   = {2nd},
  publisher = {Dover},
  address   = {Mineola, NY},
  year      = {2001}
}

@article{bazilevs2010tsplines,
  title   = {Isogeometric analysis using {T}-splines},
  author  = {Bazilevs, Yuri and Calo, Victor M. and Cottrell, John A. and Evans, John A. and Hughes, Thomas J. R. and Lipton, Scott and Scott, Michael A. and Sederberg, Thomas W.},
  journal = {Computer Methods in Applied Mechanics and Engineering},
  volume  = {199},
  number  = {5--8},
  pages   = {229--263},
  year    = {2010},
  doi     = {10.1016/j.cma.2009.02.036}
}

@inproceedings{korzun2022des,
  title     = {Application of a Detached Eddy Simulation Approach with Finite-Rate Chemistry to {Mars}-Relevant Retropropulsion Operating Environments},
  author    = {Korzun, Ashley M. and Nastac, Gabriel and Walden, Aaron and Nielsen, Eric J. and Jones, William T. and Moran, Patrick},
  booktitle = {AIAA SciTech 2022 Forum},
  number    = {AIAA 2022-2298},
  year      = {2022},
  doi       = {10.2514/6.2022-2298}
}

@book{gottlieb1977spectral,
  title     = {Numerical Analysis of Spectral Methods: Theory and Applications},
  author    = {Gottlieb, David and Orszag, Steven A.},
  series    = {CBMS-NSF Regional Conference Series in Applied Mathematics},
  volume    = {26},
  publisher = {SIAM},
  address   = {Philadelphia},
  year      = {1977}
}

@inproceedings{wu2024transolver,
  title     = {Transolver: A Fast Transformer Solver for {PDEs} on General Geometries},
  author    = {Wu, Haixu and Luo, Huakun and Wang, Haowen and Wang, Jianmin and Long, Mingsheng},
  booktitle = {International Conference on Machine Learning (ICML)},
  year      = {2024},
  note      = {arXiv:2402.02366}
}

@inproceedings{yin2023dino,
  title     = {Continuous {PDE} Dynamics Forecasting with Implicit Neural Representations},
  author    = {Yin, Yuan and Kirchmeyer, Matthieu and Franceschi, Jean-Yves and Rakotomamonjy, Alain and Gallinari, Patrick},
  booktitle = {International Conference on Learning Representations (ICLR)},
  year      = {2023},
  note      = {arXiv:2209.14855}
}

@inproceedings{serrano2023coral,
  title     = {Operator Learning with Neural Fields: Tackling {PDEs} on General Geometries},
  author    = {Serrano, Louis and Le Boudec, Lise and Kassa{\"i} Koupa{\"i}, Armand and Wang, Thomas X. and Yin, Yuan and Vittaut, Jean-No{\"e}l and Gallinari, Patrick},
  booktitle = {Advances in Neural Information Processing Systems (NeurIPS)},
  volume    = {36},
  pages     = {70581--70611},
  year      = {2023}
}

@inproceedings{chen2023insr,
  title     = {Implicit Neural Spatial Representations for Time-dependent {PDEs}},
  author    = {Chen, Honglin and Wu, Rundi and Grinspun, Eitan and Zheng, Changxi and Chen, Peter Yichen},
  booktitle = {International Conference on Machine Learning (ICML)},
  pages     = {5162--5177},
  year      = {2023}
}

@inproceedings{curvo2026mspt,
  title     = {{MSPT}: Efficient Large-Scale Physical Modeling via Parallelized Multi-Scale Attention},
  author    = {Curvo, Pedro M. P. and van de Meent, Jan-Willem and Zhdanov, Maksim},
  booktitle = {Proceedings of the IEEE/CVF Conference on Computer Vision and Pattern Recognition (CVPR)},
  pages     = {12924--12933},
  year      = {2026}
}
}

\clearpage
\maketitlesupplementary
\setcounter{table}{0}
\setcounter{figure}{0}
\setcounter{equation}{0}
\renewcommand{\thetable}{S\arabic{table}}
\renewcommand{\thefigure}{S\arabic{figure}}
\renewcommand{\theequation}{S\arabic{equation}}
\renewcommand{\theHtable}{supp.\arabic{table}}
\renewcommand{\theHfigure}{supp.\arabic{figure}}
\renewcommand{\theHequation}{supp.\arabic{equation}}
\appendix
\section{Additional operator comparison}
\label{app:operator}

Table~\ref{tab:supp-geometric-drag} compares FNO and SFNO for drag prediction after geometric warm-up.
The map is first trained for 15 epochs with the sphere and uniformity losses, followed by 100 epochs of joint training.
The map learning-rate scale multiplies the operator learning rate during joint training.
For each run, we evaluate the checkpoint with the lowest validation MSE on the same 1,154 test cars.

\begin{table}[htbp]
\centering
\small
\caption{Test $R^2$ for DrivAerNet++ drag after geometric warm-up. Map LR is relative to operator LR. SFNO at 0.03 reports the mean over seeds 1, 2 and 42. Other entries use seed 42.}
\label{tab:supp-geometric-drag}
\begin{tabular}{@{}ccc@{}}
\toprule
Map LR scale & FNO & SFNO \\
\midrule
0.10 & 0.8849 & \textbf{0.9027} \\
0.03 & 0.8898 & \textbf{0.9025} \\
\bottomrule
\end{tabular}
\end{table}

\section{Amortisation of map preparation}
\label{app:breakeven}

On DrivAerNet++, preparing the FMSNO mapping takes 10,742\,s: 8,391\,s to compute OT correspondences for all 8,119 geometries and 2,351\,s for FM pretraining using only the 5,817 training geometries.
We estimate how many predictions on new geometries recover this preparation cost through inference savings over OTNO.
Each prediction includes a new OT solve for OTNO.
For preparation time $T_{\mathrm{prep}}=10{,}742$\,s and per-geometry inference times $t_{\mathrm{FMSNO}}$ and $t_{\mathrm{OTNO}}$, the break-even count is
\begin{equation}
K=\left\lceil\frac{T_{\mathrm{prep}}}{t_{\mathrm{OTNO}}-t_{\mathrm{FMSNO}}}\right\rceil.
\end{equation}
This calculation excludes operator training for both methods and OTNO training-data preprocessing.
It measures amortisation of the FMSNO mapping preparation, rather than a comparison of total training costs.

\begin{table}[htbp]
\centering
\small
\caption{Estimated break-even count $K$ of new-geometry predictions. End-to-end inference times include OT for OTNO and were measured on one A100 80\,GB, with the same FNO grid size for both methods at each $N$. The 2,000- and 8,000-point inputs are real meshes; larger inputs are synthetic surfaces used for timing.}
\label{tab:supp-breakeven}
\setlength{\tabcolsep}{4pt}
\begin{tabular}{@{}rrrr@{}}
\toprule
Points $N$ & FMSNO (s) & OTNO (s) & $K$ \\
\midrule
2,000 & 0.02831 & 0.20829 & 59,685 \\
8,000 & 0.05713 & 0.47101 & 25,955 \\
32,000 & 0.20702 & 2.80199 & 4,140 \\
128,000 & 0.78262 & 36.86932 & 298 \\
256,000 & 1.55770 & 141.09479 & 77 \\
512,000 & 3.01956 & 555.50428 & 20 \\
1,000,000 & 6.10300 & 2162.52518 & 5 \\
\bottomrule
\end{tabular}
\end{table}

\section{Training configurations}
\label{app:hparams}

\begin{table}[htbp]
\centering
\small
\caption{Surface points per geometry before and after preprocessing. After counts are dataset means. The approximate raw count for AhmedML is representative; that for DrivAerNet++ is the mean over the test set. ShapeNet-Car is unchanged.}
\label{tab:supp-downsampling}
\setlength{\tabcolsep}{4pt}
\begin{tabular}{@{}lrrr@{}}
\toprule
Dataset & Before & After & Reduction \\
\midrule
ShapeNet-Car & 3,586 & 3,586 & $1\times$ \\
AhmedML & $\sim1.1$M & $\sim135$k & $\sim8\times$ \\
DrivAerNet++ & $\sim448$k & $\sim16.2$k & $\sim28\times$ \\
\bottomrule
\end{tabular}
\end{table}

Tables~\ref{tab:supp-map}--\ref{tab:supp-fm} give the mapping, pretraining and joint-training settings.
The pressure experiments use the same architecture and optimiser settings on ShapeNet-Car, AhmedML and DrivAerNet++.
The drag experiments use FNO or SFNO with the task-specific settings in Table~\ref{tab:supp-joint}.
Each optimisation step processes one geometry.

\begin{table}[htbp]
\centering
\small
\caption{Shared mapping and spherical-grid settings. The gradient-point limit applies to backpropagation through the mapping; the forward pass uses all input points.}
\label{tab:supp-map}
\setlength{\tabcolsep}{3pt}
\begin{tabular}{@{}p{0.59\columnwidth}p{0.35\columnwidth}@{}}
\toprule
Setting & Value \\
\midrule
Velocity-network hidden layers & 3 \\
Hidden width & 256 \\
Time-embedding dimension & 64 \\
Activation & SiLU \\
ODE solver & RK4 \\
Integration steps & 8 \\
Integration interval & $[0,1]$ \\
Spherical grid & Equiangular \\
Grid expansion factor $\alpha$ & 3 \\
Splatting and interpolation & Bilinear \\
Mapping gradient-point limit & 10,000 \\
Uniformity sample limit & 2,048 points \\
\bottomrule
\end{tabular}
\end{table}

\begin{table*}[t]
\centering
\small
\caption{Joint-training settings for FMSNO. All four variants use FM pretraining; pressure SFNO is the DrivAerNet++ operator ablation. The SFNO implementation disables domain padding.}
\label{tab:supp-joint}
\setlength{\tabcolsep}{7pt}
\begin{tabular}{@{}lcccc@{}}
\toprule
Setting & Pressure, FNO & Pressure, SFNO & Drag, FNO & Drag, SFNO \\
\midrule
Hidden channels & 120 & 120 & 120 & 120 \\
Spectral modes & $32\times32$ & $32\times32$ & $32\times32$ & $32\times32$ \\
Operator layers & 4 & 4 & 4 & 4 \\
Lifting / projection width & 256 / 256 & 256 / 256 & 256 / 256 & 256 / 256 \\
Channel-MLP expansion & 1 & 1 & 1 & 1 \\
Channel-MLP dropout & 0 & 0 & 0 & 0 \\
Normalisation & Group norm & Group norm & Group norm & Group norm \\
Domain padding & 0.125 & Disabled & 0.125 & Disabled \\
Joint epochs & 100 & 100 & 100 & 100 \\
Operator learning rate & $10^{-3}$ & $10^{-3}$ & $10^{-3}$ & $10^{-3}$ \\
Map LR scale & 0.10 & 0.10 & 0.03 & 0.03 \\
Weight decay & $10^{-4}$ & $10^{-4}$ & $10^{-4}$ & $10^{-4}$ \\
Learning-rate schedule & Cosine & Cosine & Cosine & Cosine \\
Gradient clipping norm & 1 & 1 & 5 & 5 \\
Data-loss weight & 1 & 1 & 1 & 1 \\
Sphere-loss weight & 3 & 3 & 3 & 3 \\
Uniformity-loss weight & 1 & 1 & 1 & 1 \\
\bottomrule
\end{tabular}
\end{table*}

\paragraph{Flow-matching pretraining.}
We form point pairs using the precomputed OT correspondences and sample 4,096 pairs per training geometry, with replacement, at each epoch.
For every pair, the interpolation time is sampled uniformly from $[0,1]$.
The velocity network is trained against the straight-line path velocity with mean squared error.
We use the checkpoint from epoch 200 to initialise joint training.
The FNO and SFNO drag runs use the DrivAerNet++ FM checkpoint at seed 42.

\begin{table}[htbp]
\centering
\small
\caption{Flow-matching pretraining settings. The point-pair sample is refreshed each epoch.}
\label{tab:supp-fm}
\setlength{\tabcolsep}{3pt}
\begin{tabular}{@{}p{0.59\columnwidth}p{0.35\columnwidth}@{}}
\toprule
Setting & Value \\
\midrule
Epochs & 200 \\
Pairs per geometry & 4,096 \\
Optimiser & Adam \\
Initial learning rate & $10^{-3}$ \\
Weight decay & 0 \\
Learning-rate schedule & Cosine, 200 epochs \\
Training loss & Velocity MSE \\
Checkpoint & Final epoch \\
\bottomrule
\end{tabular}
\end{table}

\paragraph{Joint training.}
Pressure values and drag targets are standardised as
\begin{equation}
\begin{aligned}
\widetilde p_i &= \frac{p_i-\mu_{p,\mathrm{train}}}{\sigma_{p,\mathrm{train}}}, \\
\widetilde C_D &= \frac{C_D-\mu_{C_D,\mathrm{train}}}{\sigma_{C_D,\mathrm{train}}},
\end{aligned}
\end{equation}
where all means and standard deviations are computed from the training set.
We select pressure checkpoints by validation relative $L_2$ error and drag checkpoints by validation MSE.
On ShapeNet-Car, the same 111 geometries are used for epoch selection and reporting, following the protocol described in the main paper.
The pressure loss is evaluated per geometry; reported pressure errors are computed in physical units and averaged over geometries.

\paragraph{Geometric warm-up.}
The pressure ablation uses 50 warm-up epochs; the additional drag comparison in Table~\ref{tab:supp-geometric-drag} uses 15.
During warm-up, we minimise
\begin{equation}
\mathcal L_{\mathrm{warmup}}=\mathcal L_{\mathrm{sph}}+\mathcal L_{\mathrm{unif}},
\end{equation}
with a mapping learning rate of $10^{-3}$.
We sample up to 10,000 surface points per geometry for this stage.
Joint training then uses the settings in Table~\ref{tab:supp-joint}, with the drag map LR scale varied as shown in Table~\ref{tab:supp-geometric-drag}.

\paragraph{Ablation settings and seeds.}
The reference pressure results are means over seeds 1, 2 and 42.
Each ablation uses seed 42; resampling and collision diagnostics also use seed 42.
We report means without error bars in the ablation tables and state the seed coverage in each caption.

\end{document}